\documentclass[10pt,twocolumn,letterpaper]{article}

\usepackage{cvpr}              

\usepackage{xcolor}
\usepackage{listings}
\lstdefinestyle{json}{
    basicstyle=\ttfamily\footnotesize,
    numbers=left,
    numberstyle=\tiny,
    stepnumber=1,
    numbersep=5pt,
    showstringspaces=false,
    breaklines=true,
    literate=
     *{0}{{{\color{blue}0}}}{1}
      {1}{{{\color{blue}1}}}{1}
      {2}{{{\color{blue}2}}}{1}
      {3}{{{\color{blue}3}}}{1}
      {4}{{{\color{blue}4}}}{1}
      {5}{{{\color{blue}5}}}{1}
      {6}{{{\color{blue}6}}}{1}
      {7}{{{\color{blue}7}}}{1}
      {8}{{{\color{blue}8}}}{1}
      {9}{{{\color{blue}9}}}{1}
      {:}{{{\color{punct}{:}}}}{1}
      {,}{{{\color{punct}{,}}}}{1}
      {\{}{{{\color{delim}{\{}}}}{1}
      {\}}{{{\color{delim}{\}}}}}{1}
      {[}{{{\color{delim}{[}}}}{1}
      {]}{{{\color{delim}{]}}}}{1},
}
\definecolor{delim}{RGB}{20,105,176}
\definecolor{punct}{RGB}{255,0,0}

\usepackage{multirow}

\usepackage{float}

\definecolor{cvprblue}{rgb}{0.21,0.49,0.74}
\usepackage[pagebackref,breaklinks,colorlinks,allcolors=cvprblue]{hyperref}

\def\paperID{*****} 
\def\confName{CVPR}
\def\confYear{2025}

\title{Instant Particle Size Distribution Measurement Using CNNs Trained on Synthetic Data}

\author{
Yasser Eljarida \quad Youssef Iraqi \quad Loubna Mekouar \\
College of Computing, University Mohammed VI Polytechnic \\
Benguerir, Morocco \\
{\tt\small \{yasser.eljarida, youssef.iraqi, loubna.mekouar\}@um6p.ma}
}

\begin{document}
\maketitle
\begin{abstract}
\textit{
Accurate particle size distribution (PSD) measurement is important in industries such as mining, pharmaceuticals, and fertilizer manufacturing, significantly influencing product quality and operational efficiency. Traditional PSD methods like sieve analysis and laser diffraction are manual, time-consuming, and limited by particle overlap. Recent developments in convolutional neural networks (CNNs) enable automated, real-time PSD estimation directly from particle images. In this work, we present a CNN-based methodology trained on realistic synthetic particle imagery generated using Blender’s advanced rendering capabilities. Synthetic data sets using this method can replicate various industrial scenarios by systematically varying particle shapes, textures, lighting, and spatial arrangements that closely resemble the actual configurations. We evaluated three CNN-based architectures—ResNet-50, InceptionV3, and EfficientNet-B0—for predicting critical PSD parameters (d10, d50, d90). Results demonstrated comparable accuracy across models, with EfficientNet-B0 achieving the best computational efficiency suitable for real-time industrial deployment. This approach shows the effectiveness of realistic synthetic data for robust CNN training, which offers significant potential for automated industrial PSD monitoring.
    }
\end{abstract}
    
\section{Introduction}
\label{sec:intro}

Accurate particle size distribution (PSD) directly impacts material properties, process efficiency, and product consistency across multiple industries, including mining, pharmaceuticals, and agriculture. Pharmaceutical manufacturing relies heavily on accurate PSD control for drug uniformity, bioavailability, and effectiveness~\cite{Kim2023}, while fertilizer production requires consistent PSD management for optimal nutrient release and spreading performance~\cite{Maria2016}. Given these implications, accurate and reliable real-time PSD measurement methods are essential.

Conventional PSD measurement approaches, such as sieve analysis and laser diffraction~\cite{Maria2016}, have significant limitations. Sieving is straightforward but struggles with accuracy, especially for fine particles~\cite{Maria2016}. Laser diffraction provides precise measurements but relies heavily on the assumption of spherical particles, limiting its effectiveness for irregular particle shapes common in practical scenarios~\cite{Capone2019}. While microscopy-based methods can deliver detailed particle morphology, they remain labor-intensive, slow, and prone to human annotation errors~\cite{Labati2019}.

Recent advancements in convolutional neural networks (CNNs) have opened new opportunities for automating PSD estimation directly from particle images~\cite{Kittiworapanya2021,Zheng2022}. Unlike traditional segmentation-based computer vision methods, CNNs are robust to overlapping particles and variations in lighting and texture, significantly reducing annotation complexity and improving accuracy. However, CNN training generally requires large, well-annotated datasets, which are expensive and challenging to obtain through manual labeling.

Synthetic data generation addresses this issue by enabling scalable creation of fully annotated datasets with precise control over particle characteristics~\cite{zhang2021,Dahmen2024}. Earlier synthetic approaches, such as discrete element method (DEM) simulations, accurately captured particle physics but often lacked visual realism, limiting model generalization~\cite{zhang2021}. Modern graphics engines like Blender~\cite{Blend18} provide sophisticated physics simulations and photorealistic rendering capabilities, enabling generation of highly realistic synthetic particle imagery. Recent studies also emphasize the significant impact of particle shape and size distribution on material properties~\cite{Gong2024}, further motivating the creation of realistic synthetic datasets.

In this work, we leverage Blender’s advanced rendering engine to generate realistic synthetic datasets closely matching actual fertilizer granulation scenarios, with a particular focus on critical PSD metrics (d10, d50, d90). We comprehensively evaluate three popular CNN architectures—ResNet-50, InceptionV3, and EfficientNet-B0—for their ability to accurately predict PSD parameters directly from these synthetic images.

Our key contributions include: (i) introducing a robust CNN-based PSD estimation method optimized for real-time industrial monitoring; (ii) developing a flexible, automated synthetic data generation pipeline using Blender, capable of replicating diverse industrial scenarios with high fidelity; (iii) providing comprehensive evaluations demonstrating CNN performance comparable to traditional PSD measurement methods; and (iv) highlighting the practical feasibility and computational efficiency of deploying CNN models, especially EfficientNet-B0, in resource-constrained environments.

The remainder of this paper is structured as follows: Section~\ref{sec:background} reviews related PSD estimation methods. Section~\ref{sec:data_gen} details our synthetic data generation pipeline. Section~\ref{sec:models} outlines CNN training and model selection strategies. Section~\ref{sec:results} presents experimental results and analyses. Section~\ref{sec:discussion} discusses practical implications and domain adaptation strategies, and Section~\ref{sec:conclusion} provides concluding remarks.

\section{Background and Related Works}
\label{sec:background}

\subsection{Classical PSD Measurement Methods}

Classical methods for measuring PSD remain widely adopted despite inherent limitations~\cite{Maria2016}. Sieve analysis is the most common method due to its simplicity and low cost, yet it struggles with accurately handling fine particles or scenarios involving significant overlap~\cite{Maria2016}. Laser diffraction provides higher accuracy and a broader measurable range but assumes spherical particle shapes, leading to inaccuracies when analyzing irregularly shaped particles~\cite{Capone2019}. Sedimentation methods and Coulter counters offer precision, particularly for fine particle ranges, but both methods require meticulous sample preparation and extensive analysis time, thus limiting their application in real-time industrial scenarios~\cite{Kim2023,Alabdali2024}.

\subsection{Computer Vision Methods for PSD Estimation}

Computer vision-based methods provide solutions to many challenges posed by classical PSD techniques, enabling automated and rapid analysis directly from images~\cite{Labati2019,Zheng2022}. Early vision-based PSD estimation relied heavily on explicit particle segmentation from images. However, these methods frequently failed under challenging conditions such as overlapping particles, varied lighting, or complex backgrounds~\cite{Maria2016}.

Recent advances using CNNs address these limitations by learning to directly estimate PSD metrics (e.g., d10, d50, d90) from raw image data without explicit segmentation~\cite{Kittiworapanya2021,Zheng2022}. CNN-based regression models have shown significant robustness to common real-world issues such as overlapping particles, irregular shapes, and variable lighting conditions~\cite{Kittiworapanya2021,Kim2023}. However, training effective CNN models typically requires large annotated datasets, which are difficult and costly to acquire from real-world scenarios~\cite{zhang2021}.

\subsection{Synthetic Data Generation for PSD}

Synthetic data generation has emerged as a practical solution to overcome the data acquisition bottleneck by allowing researchers to generate large-scale annotated datasets through computational simulations~\cite{zhang2021}. Previous synthetic generation methods, such as those based on DEM simulations, produced accurately annotated data but often lacked visual realism in particle appearance and texture~\cite{zhang2021}.

Recent progress by Dahmen et al.~\cite{Dahmen2024} highlighted the potential of neural rendering techniques for generating realistic synthetic scanning electron microscopy (SEM) images efficiently. This approach significantly improved realism while maintaining low computational costs compared to traditional methods. Motivated by these developments, our work adopts Blender, a 3D graphics engine~\cite{Blend18}, to generate synthetic particle images with detailed particle geometries, realistic textures, and highly controllable environmental parameters. By incorporating advanced rendering techniques and systematic randomization of particle properties such as size distribution and spatial arrangements, we produce synthetic datasets closely aligned with real-world particle imaging scenarios. This capability significantly enhances the effectiveness of CNN-based PSD estimation methods, enabling robust and accurate predictions in practical applications.

\section{Synthetic Data Generation}
\label{sec:data_gen}

Manually annotating large-scale datasets for PSD analysis is costly, error-prone, and challenging to scale. To overcome these limitations, we developed a fully automated synthetic data generation pipeline using Blender, an open-source 3D graphics software~\cite{Blend18}. Blender makes it possible to have detailed control over particle geometries, materials, and environmental conditions, enabling the creation of diverse and realistic particle scenes.

All the configuration values that we will present in this section are just specific to our generated dataset and can be changed to simulate any given environment from particles shape and size to the environment setup.

\subsection{Simulation Environment}

\begin{figure*}[htbp]
    \centering
    \begin{subfigure}{0.48\linewidth}
        \centering
        \includegraphics[height=5cm]{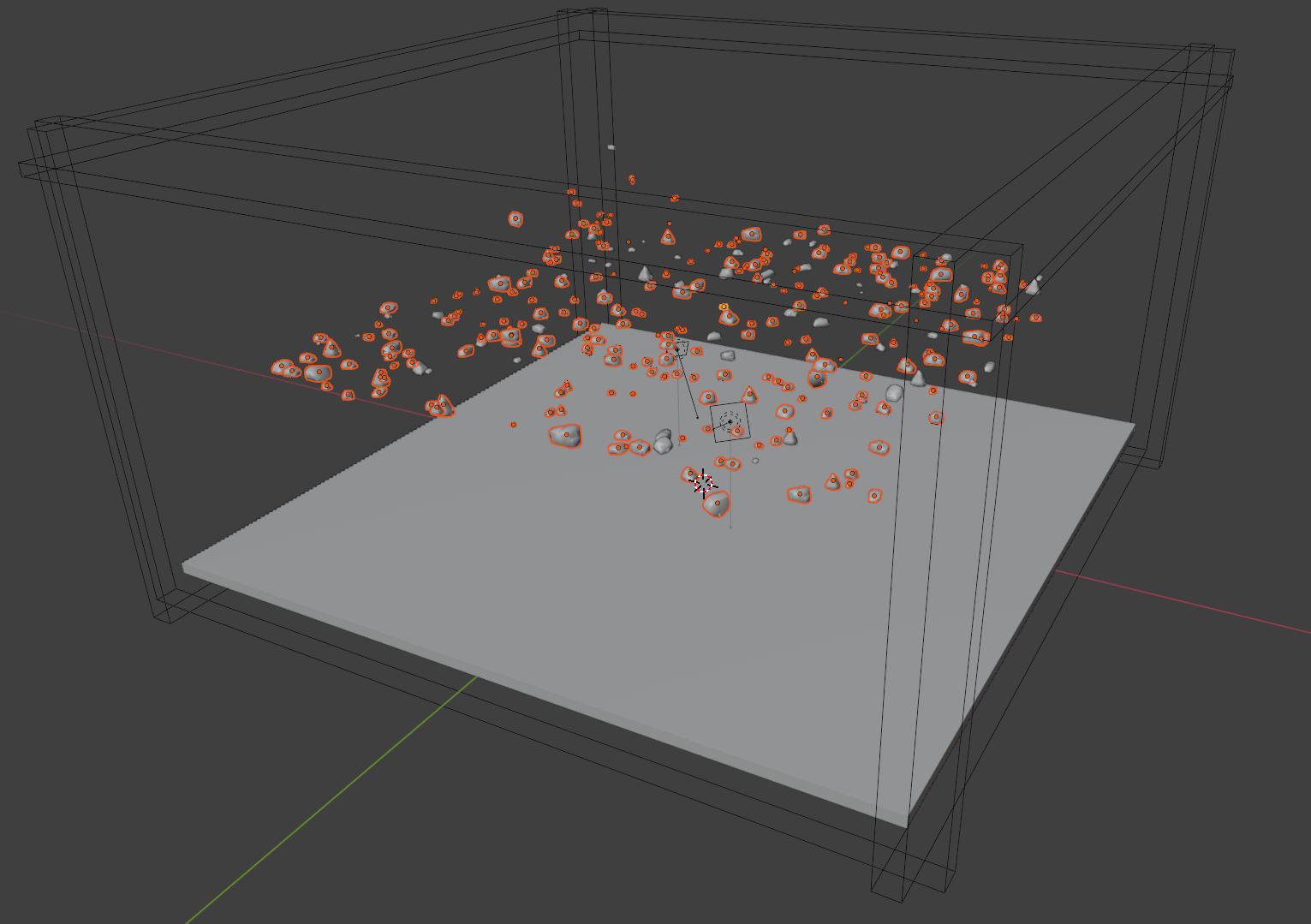}
        \caption{Setup inside the Blender interface.}
        \label{fig:blender_interface}
    \end{subfigure}
    \hfill
    \begin{subfigure}{0.48\linewidth}
        \centering
        \includegraphics[height=5cm]{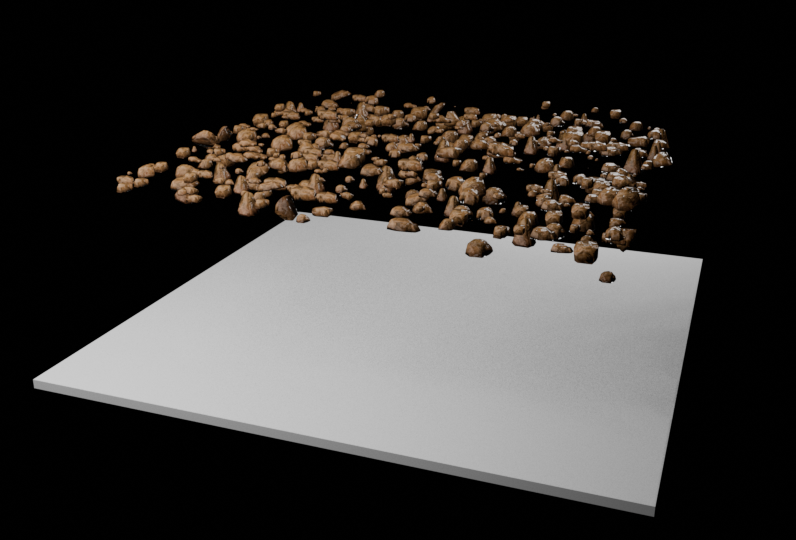}
        \caption{Rendered image depicting particle deposition before fall.}
        \label{fig:blender_rendered}
    \end{subfigure}
    \caption{Simulation environment setup in Blender.}
    \label{fig:blender_setup}
\end{figure*}

Our virtual simulation environment replicates a realistic setup for taking sample images from processes such as fertilizer granulation or aggregate handling. The scene includes a flat square surface for particle deposition measuring 300\,mm $\times$ 300\,mm, and invisible collision walls surrounding the table to ensure particles remain within the simulation area, consistent with the generated metadata (Fig.~\ref{fig:blender_setup}).

\begin{figure*}[htbp]
    \centering
    \begin{subfigure}{0.48\linewidth}
        \centering
        \includegraphics[height=4cm]{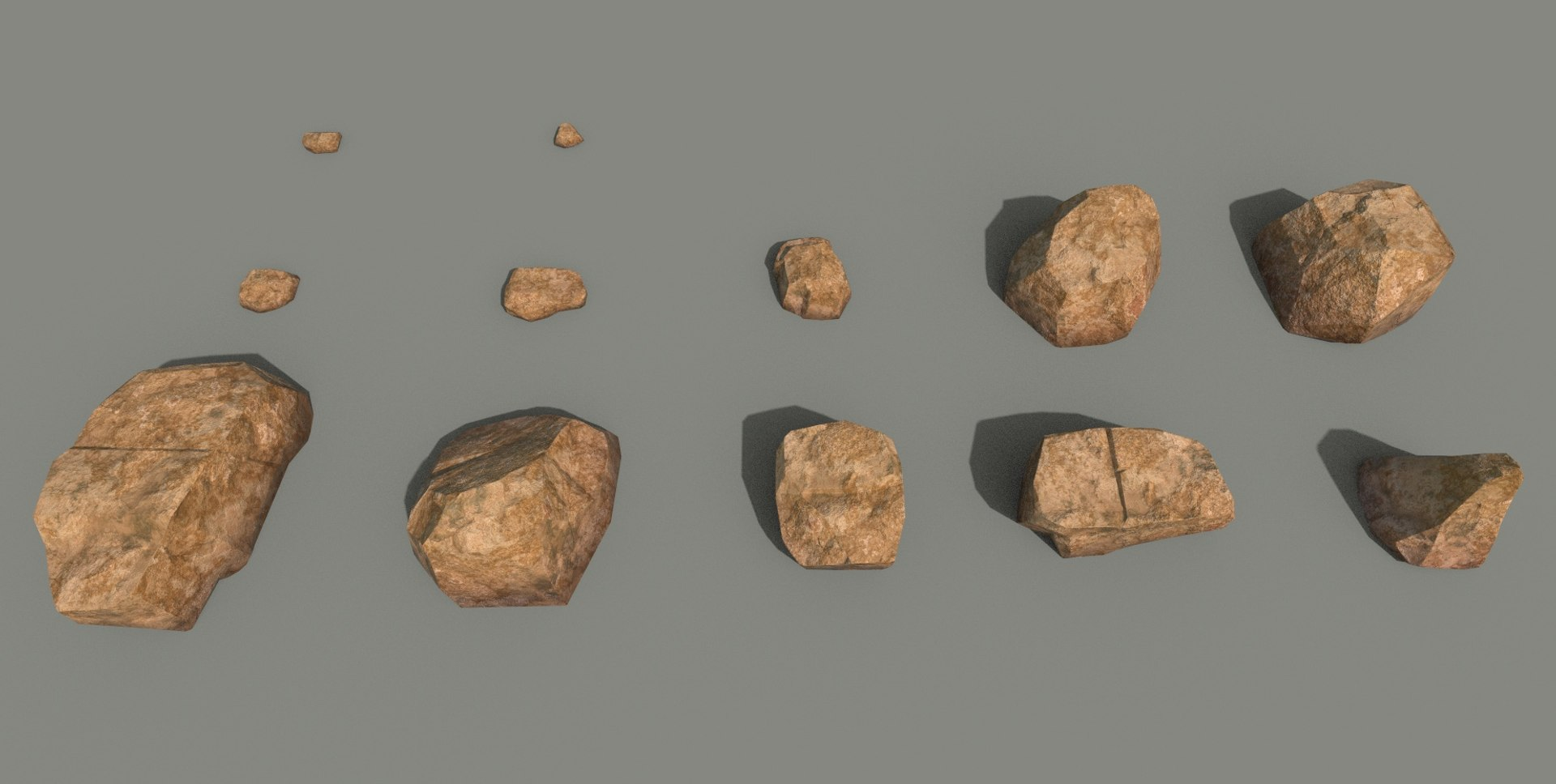}
        \caption{Rendered particle models with realistic textures.}
        \label{fig:rock_rendered}
    \end{subfigure}
    \hfill
    \begin{subfigure}{0.48\linewidth}
        \centering
        \includegraphics[height=4cm]{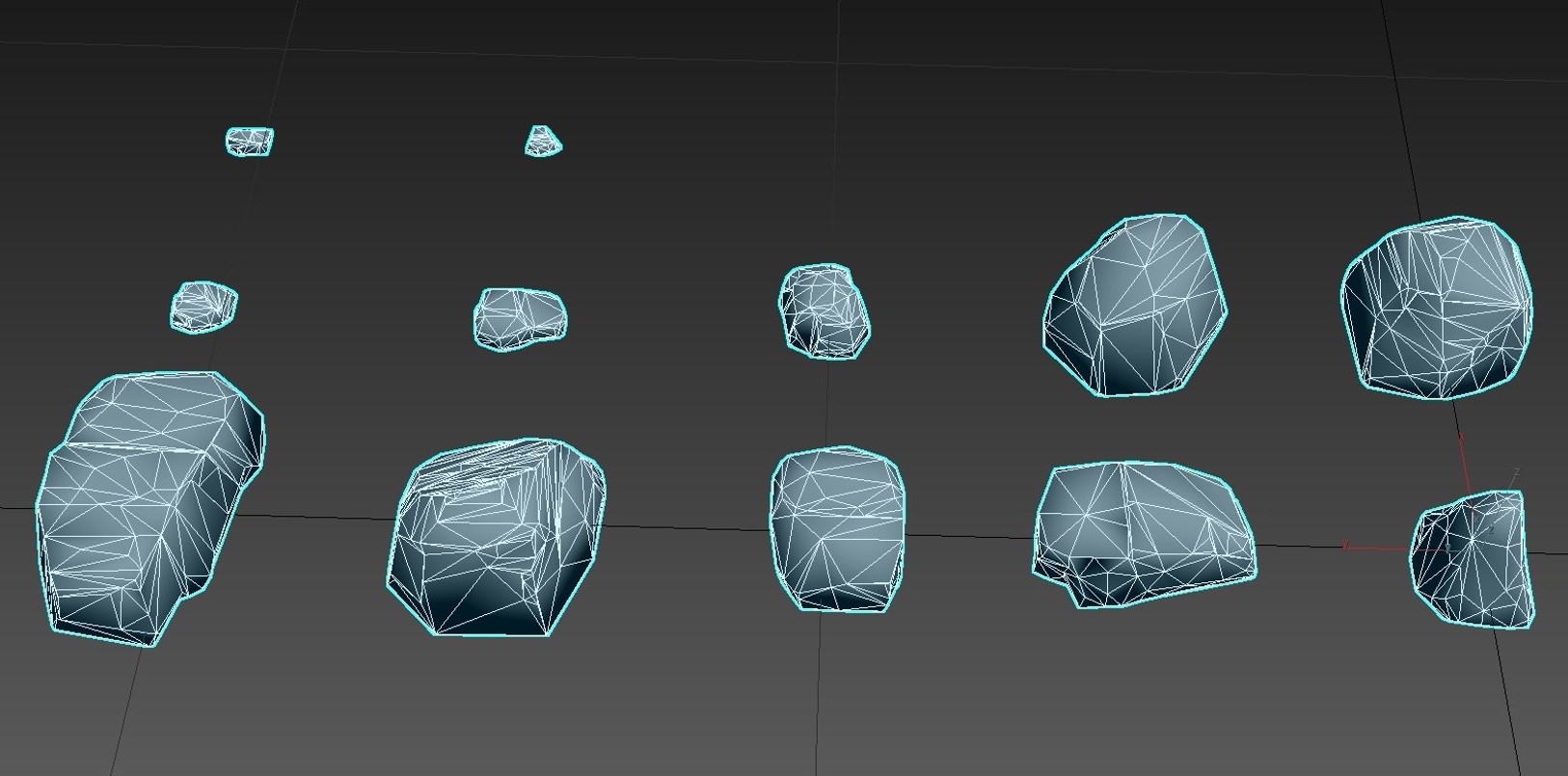}
        \caption{Wireframe mesh showing particle geometric complexity.}
        \label{fig:rock_mesh}
    \end{subfigure}
    \caption{Particle model geometry and texture realism.}
    \label{fig:rock_models}
\end{figure*}

Particles were modeled using detailed 3D meshes imported as FBX files with physically accurate textures, including base color, metallic properties, roughness, and normal maps. The complexity of these particle models, shown in Fig.~\ref{fig:rock_models}, significantly enhances the realism of the synthetic data, surpassing traditional DEM-based simulations, which typically lack such visual fidelity.

\subsection{Automated Dataset Generation Pipeline}

The dataset generation involves two interconnected Python scripts integrated within Blender’s Python API:

\begin{enumerate}
    \item \textbf{Metadata Generation}: This script randomly selects parameters for each simulation, including particle count, particle sizes, and positions. Particle sizes follow a truncated normal distribution defined as:
\begin{equation}
    \text{size} \sim \text{TruncNormal}(a, b, \mu, \sigma),
\end{equation}

    \begin{itemize}
        \item ($a, b$) : Truncation bounds.
        \item ($\mu$) : Mean particle size.
        \item ($\sigma$) : Standard deviation.
        
    \end{itemize}

For our dataset we used the configuration ranges shown in Table \ref{tab:data configuration}:

\begin{table}[H]
    \centering
    \begin{tabular}{ccc}
     & Ranges\\
    \hline
       ($\mu$)  & [6 mm, 12 mm]\\
       ($\sigma$)  & [6 mm, 8 mm] \\
       ($a, b$)  & [0.1 mm, 20 mm] \\
       Particle count  & [700, 1000] \\
       \hline
    \end{tabular}
    \caption{Data configuration}
    \label{tab:data configuration}
\end{table}

The metadata for each simulation are stored as structured JSON files containing the following information: shape type, mean and standard deviation of particle sizes, table dimensions, particle count, and a list of individual particle parameters (size and x-y coordinates). Listing \ref{lst:json-example} shows an example of a sample metadata file.

\begin{lstlisting}[style=json, caption={Example JSON metadata describing a synthetic particle scene.}, label={lst:json-example}]
{
  "shape_type": "crushed_rock",
  "size_mean": 10.5,
  "size_sigma": 7.2,
  "table_size": 300,
  "samplesize": 920,
  "particles": [
    {"size": 11.80, "x": -11.12, "y": -2.31},
    {"size": 7.97, "x": -118.32, "y": 119.38},
    ...
  ]
}
\end{lstlisting}

\item \textbf{Scene Construction and Rendering}: The second script reads the generated JSON metadata, places particles accordingly, and initiates Blender’s rigid-body physics simulation. Once particles settle, the final scene state is rendered using Blender's Cycles rendering engine with GPU (CUDA) acceleration, significantly enhancing rendering efficiency. The rendered images accurately reflect their corresponding metadata annotations.

\end{enumerate}

This pipeline was executed on a High Performance Computing (HPC) system using NVIDIA A100 GPUs, facilitating rapid and scalable dataset generation. For our training dataset, we rendered approximately 4500 images (4668 images across three rendering jobs), each taking in average one minute to render, depending on the sample size.
\subsection{Dataset Realism and Diversity}

\begin{figure*}[htbp]
    \centering
    \includegraphics[width=12cm]{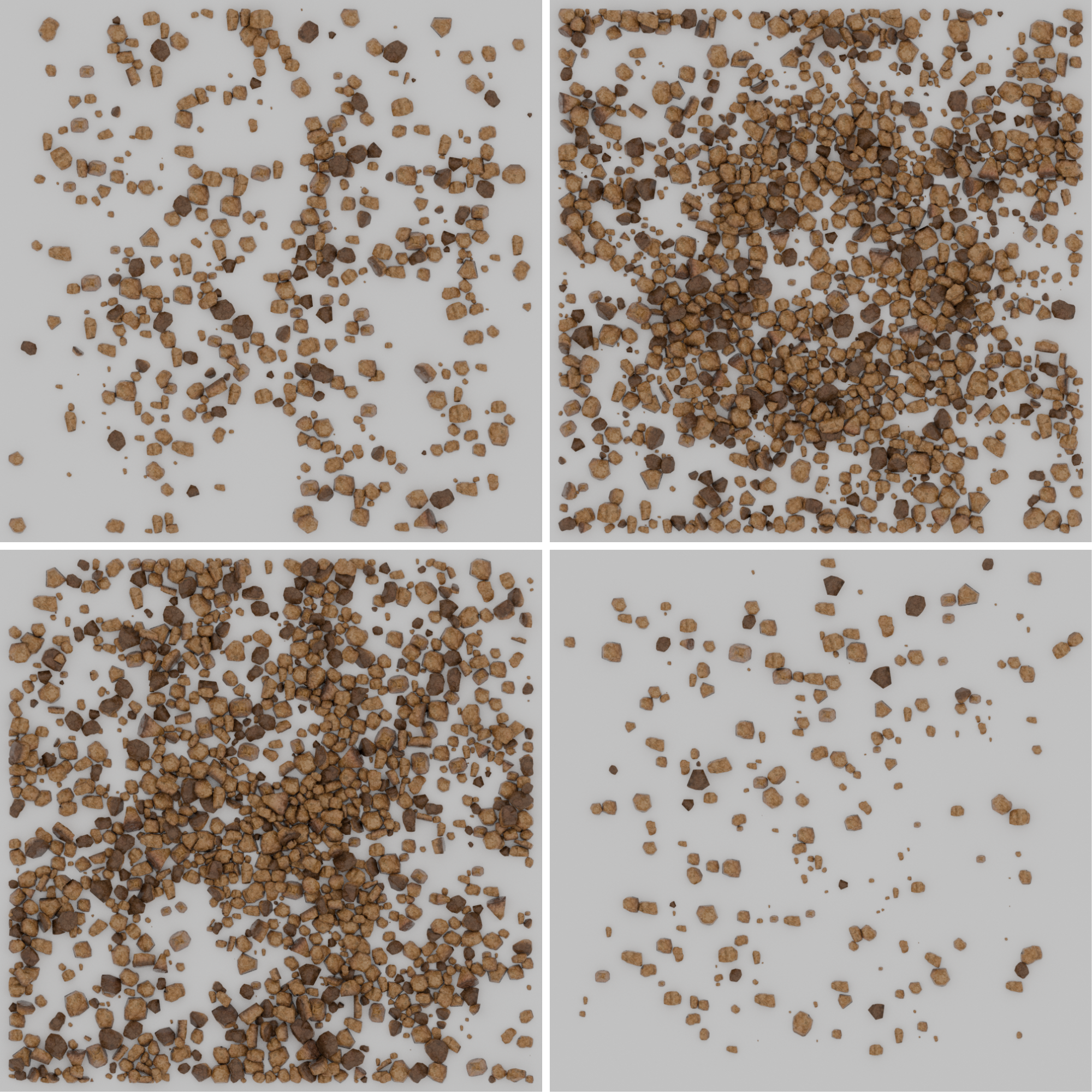}
    \caption{Examples of synthetic images demonstrating realistic variability in particle sizes, shapes, textures, and spatial distributions.}
    \label{fig:render-example}
\end{figure*}

Our synthetic images exhibit realistic lighting, detailed textures, and diverse particle arrangements (Fig.~\ref{fig:render-example}). Compared to previous DEM-based synthetic datasets~\cite{zhang2021}, our Blender-based dataset achieves significantly greater visual realism, closely approximating real-world scenarios and thus enhancing CNN model generalizability.

\subsection{Advantages over Previous Methods}

Our Blender-based synthetic data generation method offers several key advantages:

\begin{itemize}
    \item \textit{Enhanced Realism}: Complex particle geometries and detailed textures produce visually realistic datasets.
    \item \textit{Flexibility and Control}: Adjustable parameters (lighting, particle properties, environment) enable replication of a broad range of real-world conditions.
    \item \textit{Scalable Automation}: GPU-accelerated rendering combined with automated metadata handling allows efficient large-scale dataset creation.
    \item \textit{Accurate Annotation}: Automated metadata generation ensures consistent and precise annotations, eliminating manual labeling errors.
\end{itemize}

This pipeline efficiently generates realistic synthetic datasets essential for training robust and reliable CNN models for practical PSD estimation tasks.

\section{CNN Model Selection and Training}
\label{sec:models}

To evaluate the effectiveness of our synthetic dataset for PSD estimation, we selected CNNs known for strong performance, computational efficiency, and suitability for regression tasks. Our primary goal was to accurately predict critical PSD parameters (d10, d50, d90) directly from particle images.

\subsection{Model Selection}

We selected three widely-adopted CNN architectures, each chosen for specific strengths relevant to particle analysis:

\begin{itemize}
    \item \textit{ResNet-50}~\cite{He2016ResNet}: Chosen for its robust feature extraction capabilities, using residual connections to effectively learn complex visual patterns without suffering from vanishing gradients.
    \item \textit{EfficientNet-B0}~\cite{Tan2019EfficientNet}: Selected because of its balanced trade-off between computational efficiency and predictive accuracy, making it particularly suitable for real-time industrial monitoring scenarios.
    \item \textit{InceptionV3}~\cite{Szegedy2016InceptionV3}: Included for its ability to extract multi-scale features via parallel convolutional operations, effectively capturing variations in particle sizes, shapes, and textures.

\end{itemize}

\subsection{Training Procedure}

All models were pretrained on the ImageNet dataset, as preliminary experiments demonstrated improved convergence speed and predictive accuracy compared to training from scratch. 

The models were fine-tuned to perform regression directly on the PSD parameters (d10, d50, d90). After some experimentation, we ended up using these Training hyperparameters:

\begin{itemize}
    \item \textit{Loss Function}: Mean Squared Error (MSE).
    \item \textit{Optimizer}: Adam optimizer with an initial learning rate of \(1 \times 10^{-4}\), adaptively reduced with respect to the validation plateau.
    \item \textit{Data Augmentation}: Random rotations (0°, 90°, 180°, 270°), horizontal flipping were employed to improve generalization and mitigate overfitting.
    \item \textit{Epochs}: Training ran up to 50 epochs which showed satisfactory convergence.
\end{itemize}

TensorBoard was utilized throughout training to monitor convergence, facilitating real-time hyperparameter tuning and performance analysis.

\subsection{Computational Resources}

Model training was conducted on an HPC cluster. Specifically, GPU-accelerated training was performed on nodes equipped with NVIDIA A100 GPUs, enabling rapid iteration, efficient experimentation, and significant reductions in overall training duration which took approximately 3 hours for each model.

The effectiveness of this model selection and training approach is evaluated in detail in the following experimental section.

\section{Experimental Results and Analysis}
\label{sec:results}

In this section, we compared the predictive accuracy and computational efficiency between the three CNN architectures (ResNet-50, InceptionV3, and EfficientNet-B0) using our synthetic dataset comprising approximately 5,000 annotated images. The evaluation focused on the prediction of key PSD parameters: d10, d50, and d90.

The performance of the models was quantitatively assessed using the coefficient of determination ($R^2$), mean squared error (MSE), and mean absolute error (MAE), providing robust measures of accuracy and consistency.

\subsection{Detailed Quantitative Results}

Table~\ref{tab:detailed_results} presents detailed predictive performance metrics (R², MSE, MAE) for each CNN model across individual PSD parameters and overall. Although ResNet-50 showed marginally better predictive performance, the differences between the three models' results are minor (less than $\approx 0.1\,\text{mm}$ error), indicating similar predictive capabilities.

\begin{table*}[htbp]
\centering
\caption{Detailed performance comparison for CNN models on synthetic test data (GPU inference). The best results are in Bold.}
\label{tab:detailed_results}
\resizebox{0.9\textwidth}{!}{
\begin{tabular}{l|ccc|ccc|ccc|ccc}
\hline
\multirow{2}{*}{CNN Model} & \multicolumn{3}{c|}{Overall} & \multicolumn{3}{c|}{d10} & \multicolumn{3}{c|}{d50} & \multicolumn{3}{c}{d90} \\ \cline{2-13} 
                           & $R^2$ & MSE   & MAE   & $R^2$ & MSE   & MAE   & $R^2$ & MSE   & MAE   & $R^2$ & MSE   & MAE   \\ \hline
ResNet-50                  & \textbf{0.9987} & \textbf{0.0204} & \textbf{0.0773} & \textbf{0.9591} & \textbf{0.0315} & \textbf{0.0913} & \textbf{0.9904} & \textbf{0.0160} & \textbf{0.0747} & \textbf{0.9888} & \textbf{0.0137} & \textbf{0.0660} \\
InceptionV3                & 0.9966 & 0.0550 & 0.1796 & 0.8933 & 0.0821 & 0.2212 & 0.9738 & 0.0438 & 0.1651 & 0.9680 & 0.0391 & 0.1525 \\
EfficientNet-B0            & 0.9959 & 0.0660 & 0.1973 & 0.8777 & 0.0941 & 0.2357 & 0.9667 & 0.0555 & 0.1841 & 0.9604 & 0.0484 & 0.1721 \\ \hline
\end{tabular}}
\end{table*}

\begin{figure}[htbp]
\centering
\includegraphics[width=\columnwidth]{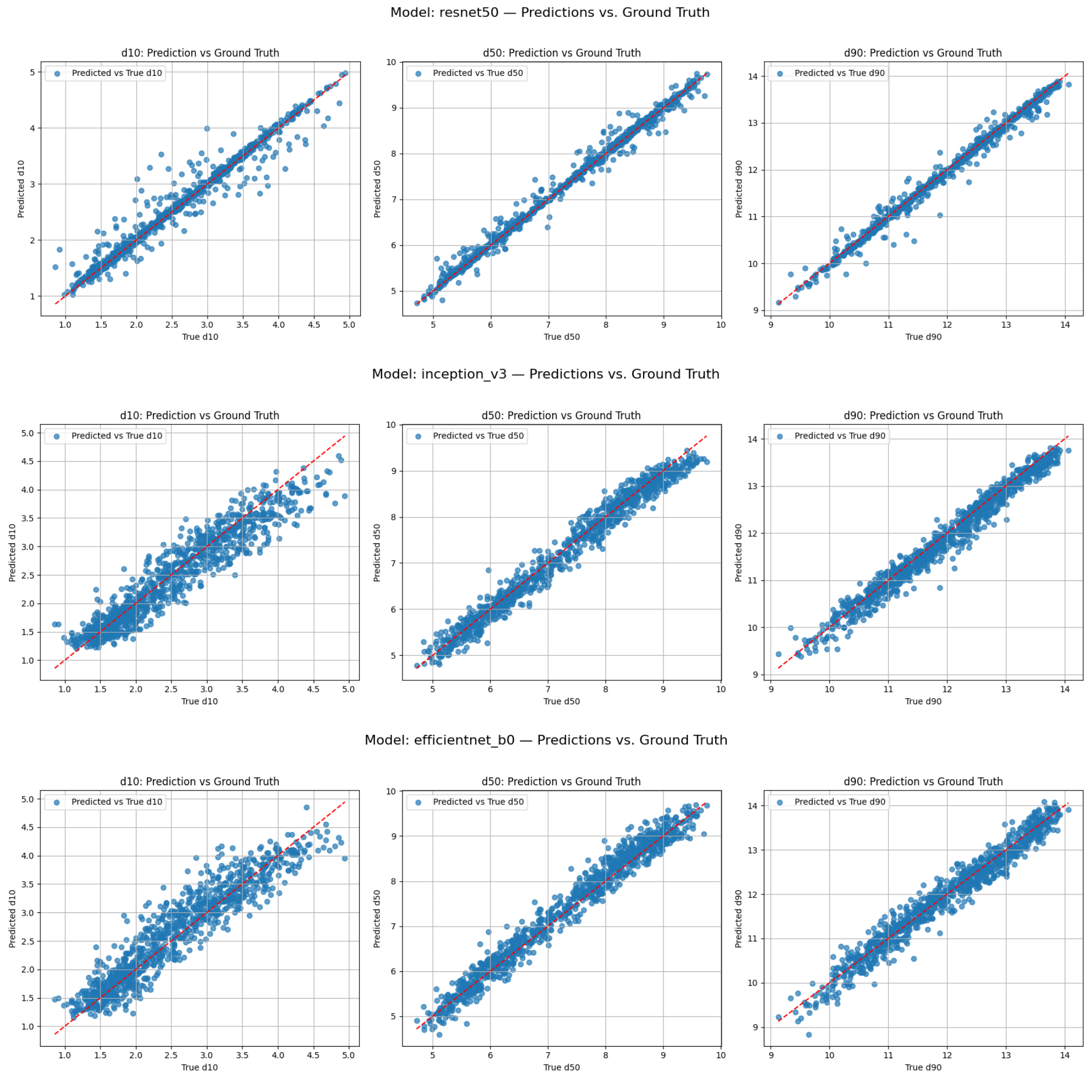}
\caption{Predicted vs. ground truth PSD comparison.}
\label{fig:pred_vs_truth}
\end{figure}

Figure~\ref{fig:pred_vs_truth} visualizes the predicted versus ground truth PSD values for each model, we can see that the ResNet-50 model is more aligned, with minor variations in prediction quality with the other models.

\begin{figure*}[htbp]
    \centering
    \begin{subfigure}{0.32\linewidth}
        \centering
        \includegraphics[height=4.2cm]{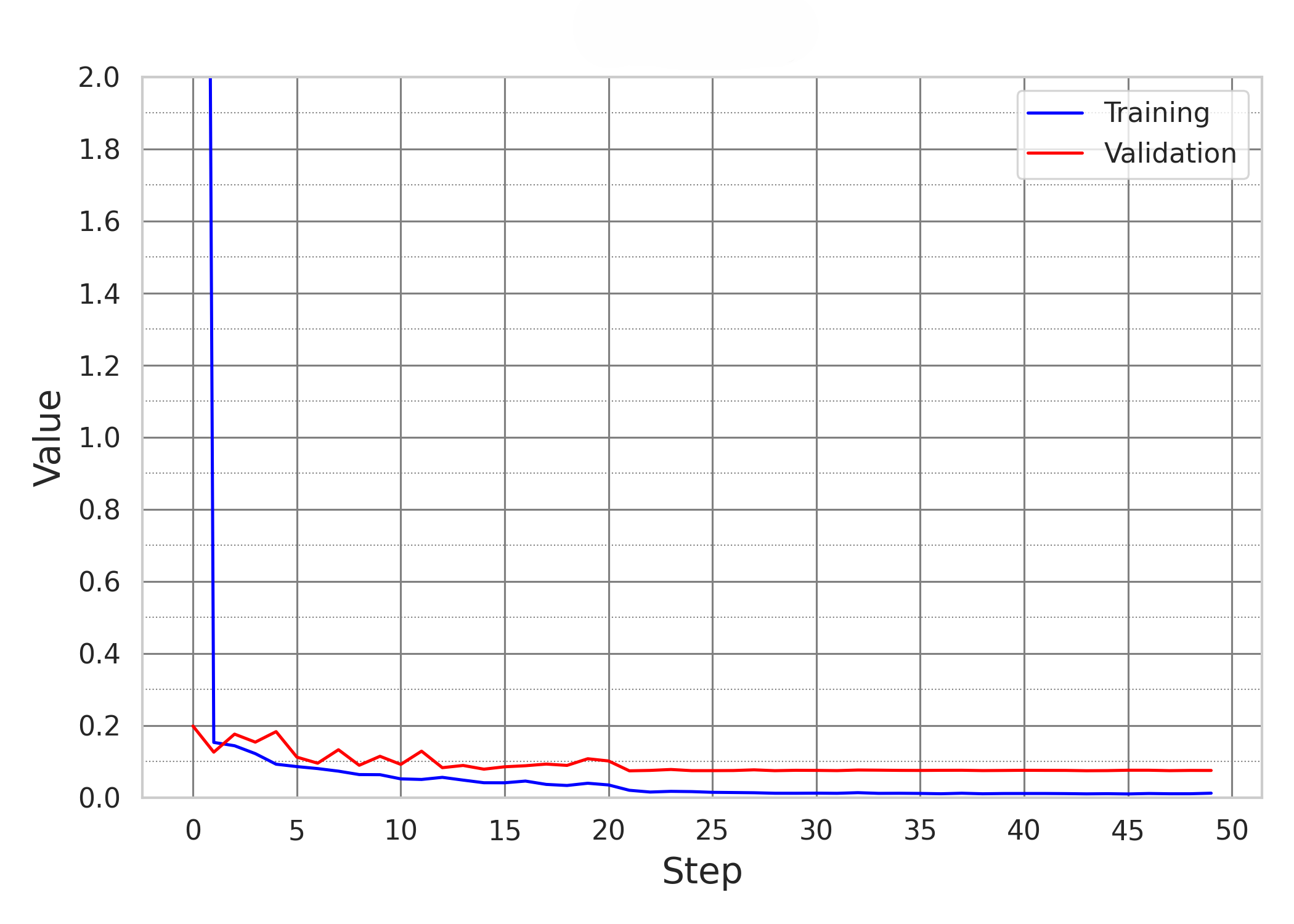}
        \caption{ResNet50}
        \label{fig:resnet50}
    \end{subfigure}
    \hfill
    \begin{subfigure}{0.32\linewidth}
        \centering
        \includegraphics[height=4.2cm]{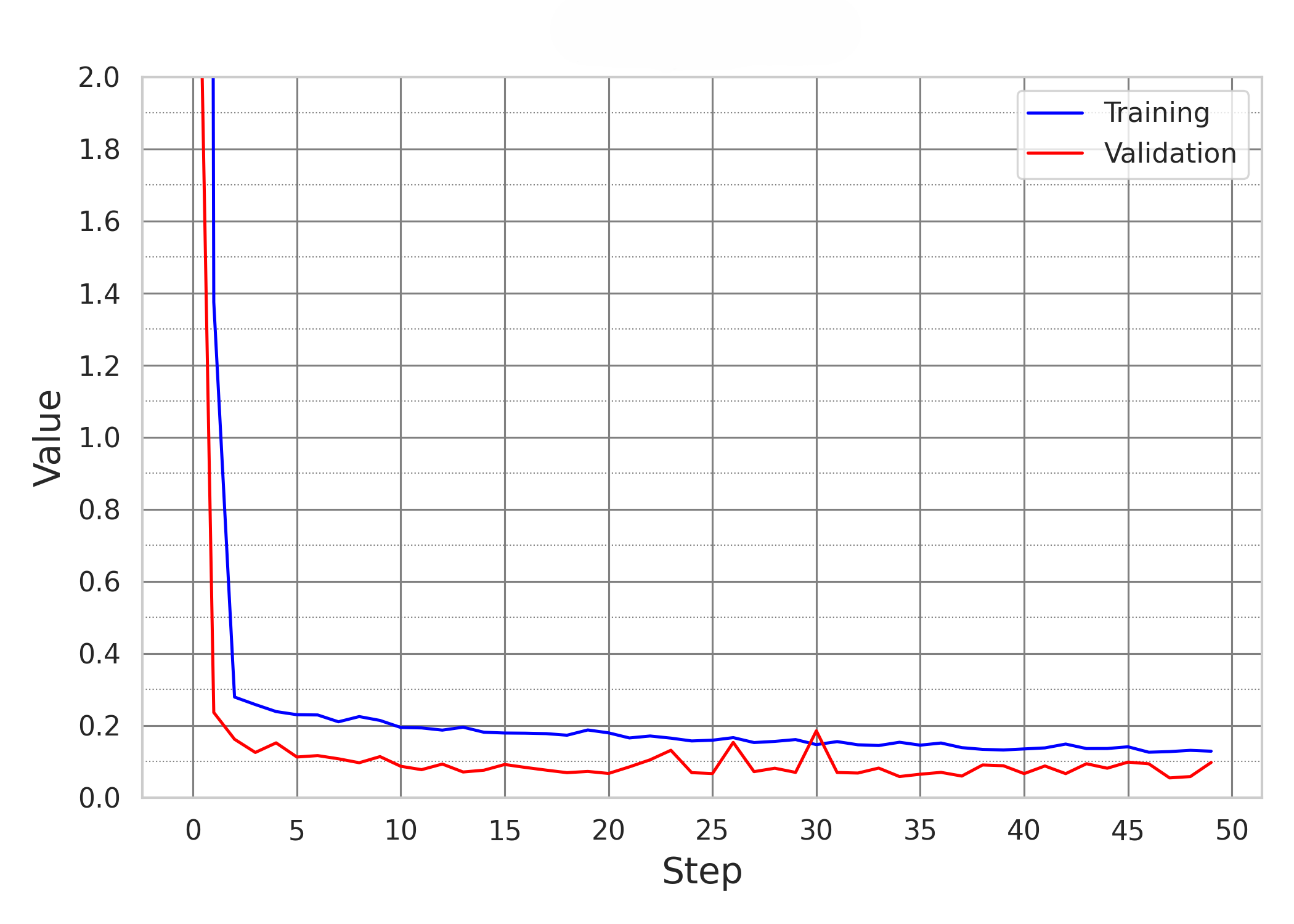}
        \caption{Inception V3}
        \label{fig:inceptionv3}
    \end{subfigure}
    \hfill
    \begin{subfigure}{0.32\linewidth}
        \centering
        \includegraphics[height=4.2cm]{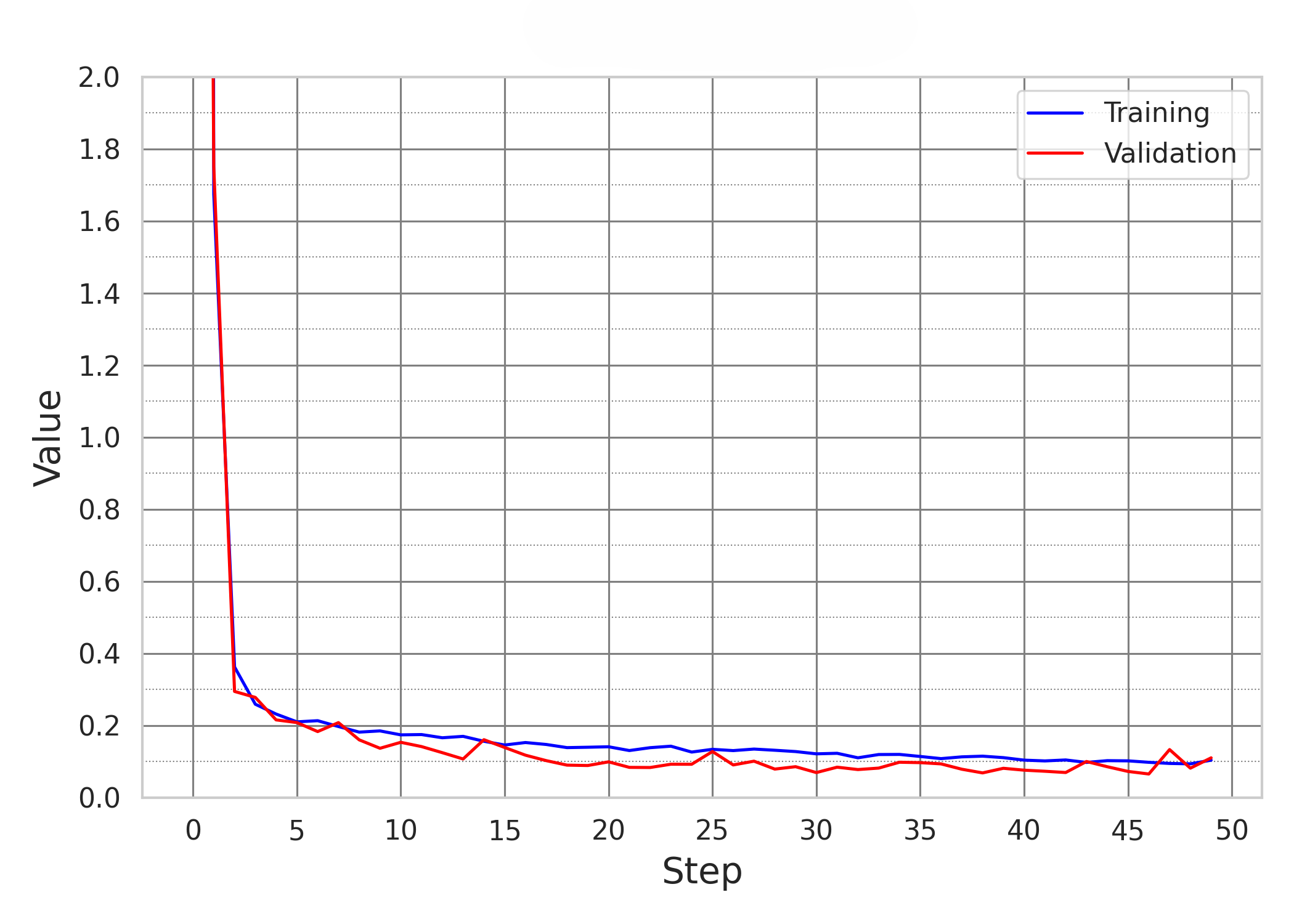}
        \caption{EfficientNet-B0}
        \label{fig:efficientnetb0}
    \end{subfigure}
    \caption{Training and validation loss curves showing model convergence for (a) ResNet50, (b) Inception V3, and (c) EfficientNet-B0.}
    \label{fig:model_loss_curves}
\end{figure*}

Figure~\ref{fig:model_loss_curves} shows training and validation loss curves monitored via TensorBoard. EfficientNet-B0 demonstrates efficient and rapid convergence behavior, which got close final validation loss to ResNet-50 and InceptionV3.

\subsection{Computational Efficiency}

Given the importance of computational speed for real-time PSD monitoring in industrial settings, we tested the inference speed of each CNN model under realistic hardware conditions. We conducted inference tests using a consumer-grade laptop equipped with an NVIDIA RTX 3060 GPU and an AMD Ryzen 7 5800H CPU, providing a practical and accessible benchmark scenario.

Inference speeds are summarized in Table~\ref{tab:efficiency}. Although ResNet-50 showed slightly better accuracy, EfficientNet-B0 significantly outperformed other models in inference speed on both GPU and CPU. Specifically, EfficientNet-B0 can process approximately 37.75 frames per second (FPS) on GPU and 21.51 FPS on CPU, markedly faster compared to ResNet-50 (27.31 FPS GPU, 14.66 FPS CPU) and InceptionV3 (23.18 FPS GPU, 12.00 FPS CPU). This substantial improvement in inference speed, combined with its considerably lower parameter count, underscores EfficientNet-B0's practical advantage, particularly in resource-constrained environments.

Thus, despite minor accuracy trade-offs, EfficientNet-B0 offers the optimal balance between accuracy and computational efficiency, making it particularly suitable for real-time PSD estimation tasks in industrial applications.

\begin{table}[htbp]
\centering
\caption{Inference speed comparison and model complexity on realistic laptop hardware (RTX 3060 GPU, AMD Ryzen 7 5800H CPU). The best results are in Bold.}
\label{tab:efficiency}
\resizebox{\columnwidth}{!}{
\begin{tabular}{lcccc}
\hline
\textbf{CNN Model} & \textbf{GPU FPS} & \textbf{CPU FPS} & \textbf{GPU Time/Image (s)} & \textbf{Parameters (M)} \\ 
\hline
ResNet-50          & 27.31   & 14.66   & 0.0366             & 25.6            \\
InceptionV3        & 23.18   & 12.00   & 0.0431             & 23.9            \\
EfficientNet-B0    & \textbf{37.75}   & \textbf{21.51}   & \textbf{0.0265}  & \textbf{5.3}   \\ 
\hline
\end{tabular}}
\end{table}

\section{Discussion}
\label{sec:discussion}

This study demonstrates the potential of using CNNs trained on realistic synthetic data generated via Blender to accurately estimate PSD. Our approach leverages Blender's advanced rendering capabilities, allowing precise control over particle geometries, textures, spatial arrangements, and lighting conditions. This meticulous control enables the creation of synthetic images that closely match real-world industrial scenarios, thereby significantly improving model robustness compared to previous approaches relying on simplified particle representations.

One key advantage of our method is the ease with which diverse scenarios can be generated. By adjusting parameters such as particle shape, size distribution (through truncated normal distributions), lighting intensity, and camera perspectives, the synthetic data pipeline becomes highly adaptable to various industrial processes or materials. Additionally, automation within Blender facilitates rapid, scalable, and cost-effective dataset generation, removing the reliance on labor-intensive manual annotations typically required for training robust CNN models.

In evaluating three popular CNN architectures—ResNet-50, InceptionV3, and EfficientNet-B0—we found minor differences in predictive performance. Although ResNet-50 achieved slightly better numerical results, EfficientNet-B0 stood out in computational efficiency, offering substantially faster inference speeds, particularly on CPU hardware. This computational advantage suggests that EfficientNet-B0 provides the most practical balance between accuracy and real-time processing, making it highly suitable for deployment in resource-constrained industrial environments or edge devices.

The realism of our synthetic dataset substantially narrows the traditional "domain gap" between synthetic and real-world imagery, yet minor discrepancies inevitably remain. Integrating targeted domain adaptation methods, such as incremental fine-tuning on limited real data or employing adversarial training techniques, offers promising ways to further enhance model generalization. These strategies can be seamlessly incorporated into the synthetic data pipeline, helping to continuously refine and align synthetic scenarios more closely with real-world operational conditions.

In practical terms, while high-performance computing resources such as GPU clusters accelerate synthetic dataset creation, our methodology does not inherently require such infrastructure. The pipeline remains operationally feasible on standard GPU-equipped workstations, but with longer rendering times. which means that computational resources should not pose a barrier to adopting our approach broadly, although they directly influence dataset generation speed.

To encourage trust in industrial deployment, we suggest initially validating CNN models alongside conventional PSD measurement techniques. This hybrid approach allows comprehensive verification and fosters confidence in automated systems, paving the way for full integration into real-time process control loops. Further enhancing model interpretability, such as employing explainability techniques (e.g., saliency maps), can also support industrial adoption by increasing transparency in model predictions.

Ultimately, the flexibility, realism, and automation offered by synthetic data generation position CNN-based PSD estimation as a highly promising alternative to traditional measurement methods, presenting substantial opportunities for improving quality control, optimizing production efficiency, and reducing operational costs across various industries.

\section{Conclusion}
\label{sec:conclusion}

This study introduced an efficient and accurate approach to the estimation of PSD in real time using CNNs trained exclusively on synthetic datasets. Using Blender's advanced rendering capabilities, we generated highly realistic particle imagery, addressing critical limitations associated with traditional PSD measurement techniques, such as manual sampling, offline analysis, and challenges related to overlapping particles.

Experimental evaluations demonstrated that CNN architectures—ResNet-50, InceptionV3, and EfficientNet-B0—can effectively predict PSD parameters (d10, d50, and d90) directly from images. Among these models, EfficientNet-B0 emerged as particularly well suited for industrial deployment due to its superior inference speed, especially on resource-constrained hardware, despite marginal and statistically insignificant accuracy differences compared to ResNet-50.

The flexibility of the synthetic data generation pipeline allows rapid adaptation to various industrial scenarios by simply adjusting particle properties, textures, and environmental settings, significantly improving training robustness and reducing data acquisition costs.

Further integration of targeted domain adaptation strategies, incremental fine-tuning on limited real-world data, and enhanced model interpretability techniques hold promise for further narrowing the synthetic-to-real gap. In general, the proposed method offers a practical, scalable, and cost-effective path to automated PSD monitoring, improving process control and product consistency in various industrial applications.

{
    \small
    \bibliographystyle{ieeenat_fullname}
    \bibliography{main}
}


\end{document}